%% file: acl_latex.tex
\documentclass[11pt]{article}

\usepackage[final]{acl}

\usepackage{times}
\usepackage{latexsym}
\usepackage{tcolorbox}
\usepackage{amsmath}
\usepackage{booktabs}
\usepackage{float}
\usepackage{amssymb}
\usepackage{microtype}
\usepackage{afterpage}
\usepackage[T1]{fontenc}
\usepackage[utf8]{inputenc}

\usepackage{microtype}

\usepackage{inconsolata}

\usepackage{graphicx}

\usepackage{booktabs}
\tcbuselibrary{breakable}

\title{Cards Against LLMs: Benchmarking Humor Alignment in Large Language Models}

\author{
  \textbf{Yousra Fettach\textsuperscript{1}},
  \textbf{Guillaume Bied\textsuperscript{1}},
  \textbf{Hannu Toivonen\textsuperscript{2}},
  \textbf{Tijl De Bie\textsuperscript{1}}
\\
\\
  \textsuperscript{1} Ghent University, Belgium \\
  \textsuperscript{2}University of Helsinki, Finland
\\
}

\begin{document}
\maketitle
%\begin{abstract}
%\yousra{Need to modify this}
%Evaluating large language models on subjective, culturally embedded tasks remains an open challenge. We introduce a large-scale humor alignment benchmark in which five Large Language Models (LLMs) compete at Cards Against Humanity alongside human players, with human game winners as ground truth. Across 9,894 rounds, no model reliably matches human humor judgment, yet inter-model agreement suggests the emergence of a collective LLM humor aesthetic distinct from human preference. 
%\end{abstract}

\begin{abstract}
Humor is one of the most culturally embedded and socially significant dimensions of human communication, yet it remains largely unexplored as a dimension of Large Language Model (LLM) alignment. In this study, five frontier language models play the same \textit{Cards Against Humanity} games (CAH) as human players. The models select the funniest response from a slate of ten candidate cards across 9,894 rounds. While all models exceed the random baseline, alignment with human preference remains modest. More striking is that models agree with each other substantially more often than they agree with humans. We show that this preference is partly explained by systematic position biases and content preferences, raising the question whether LLM humor judgment reflects genuine preference or structural artifacts of inference and alignment.
\end{abstract}

\section{Introduction}

%Large Language Models (LLMs) had made significant strides in coding \cite{}, writing and even. However, \textit{humor} remains a distinct Turing-complete challenge. Unlike logic, which converges on a single truth, humor is divergent, subjective, and deeply rooted in cultural context.

Large language models (LLMs) have made significant strides in coding \cite{chen2021evaluating}, writing \cite{brown2020language}, and creative tasks such as storytelling and poetry generation \cite{chakrabarty2022help}. This is largely attributed to training on vast human-generated corpora, emergent capabilities unlocked by scale \cite{wei2022emergent}, and reinforcement learning from human feedback (RLHF) \cite{ouyang2022training}. As these models approach human-level performance on structured tasks, attention has turned to whether they can match humans on dimensions of communication that are less formal and more culturally loaded.

Humor embodies one such dimension. It operates at the intersection of language, cognition, and social context \cite{martin2018psychology}. It also relies on the violation and resolution of expectations \cite{attardo1997semantic}, shared cultural knowledge, and sensitivity to timing, tone, and audience \cite{hay2001pragmatics}. These properties make it simultaneously deeply human and notoriously difficult to formalize \cite{mihalcea2005making}. 

%Additionally, the social role of humor is well-documented: it signals in-group membership, builds rapport, and negotiates status \cite{}. Such functions are central to any system capable of genuine human-level interaction.
This complexity makes humor a uniquely revealing test for LLMs. What a model finds funny reflects the cultural references it recognizes, the social boundaries it is willing to cross, and the moral and political values it has absorbed through training and alignment. Humor is therefore not only a capability to be scored but more of a window into the worldview of the model itself, and one of the final frontiers for AI alignment \cite{zhou2025bridging}. Probing this requires a setting that elicits genuine humor judgment while remaining structured enough for systematic comparison.

\begin{figure}%[h]
    \centering
\includegraphics[width=\linewidth]{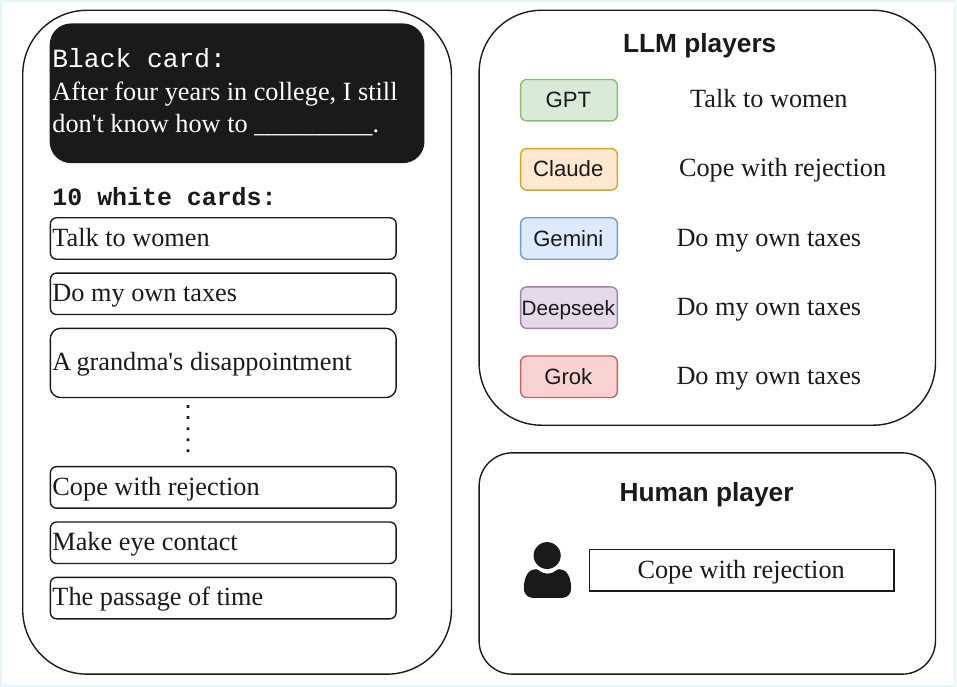}
    \caption{Framework overview. Given a black card prompt and a slate of 10 white card candidates, five frontier LLMs and a human player independently select the card they deem funniest.
    }
    \label{fig:framework}
\end{figure}

We study the alignment of LLMs with human humor through a controlled, game-based evaluation, operationally measuring it by comparing LLM-derived to human-derived preference judgments. \textit{Cards Against Humanity} (CAH) offers a natural testbed for this inquiry. As a fill-in-the-blank party game built around subversive, context-sensitive punchlines, it requires players to select the funniest response to a prompt from a fixed set of options. This makes it specific enough for systematic evaluation but also rich enough to expose meaningful variation in humor perception. We show our experimental setup in Figure ~\ref{fig:framework}. The following is a summary of our work and contributions \footnote{Code available at: \url{https://github.com/aida-ugent/cards_against_llms}}:

\begin{enumerate}
    \item We present a framework in which five frontier models, GPT-5.2, Gemini 3 Flash, Claude Opus 4.5, Grok 4, and DeepSeek-V3.2, play the same CAH games that humans have played. 
    
    \item We assess human-LLM humor alignment and find that all LLMs evaluated only achieve modest accuracy.

    \item We study the agreement between LLMs and their self-consistency across repeated runs. We demonstrate that LLMs agree with each other substantially more than they agree with humans, suggesting the emergence of stable but human-misaligned humor profiles.

    %, LLM-LLM agreement, and effects of topic choice and position bias.
%    \item On the human humor alignment evaluation, we find that all LLMs evaluated exceed the random baseline but only achieve modest accuracy.

%\item 
%We demonstrate that LLMs agree with each other substantially more than they agree with humans, suggesting the emergence of stable but human-misaligned humor profiles.

\item We provide an analysis of LLM humor selection, identifying significant position bias and specific content preferences across all five LLMs, and show that these two factors jointly explain a substantial share of inter-model agreement.
\end{enumerate}

\section{Related Work}

%\paragraph{Theories of Humour} \yousra{Revisit} Theories of humor broadly converge on three mechanisms: incongruity resolution, whereby amusement arises from the detection and reconciliation of an unexpected twist~\cite{}; superiority theory, which frames humor as a response to the perceived misfortune or inferiority of others~\cite{}; and relief theory, which views laughter as a release of psychological tension~\cite{}. 
\paragraph{LLMs and Humour Understanding} Humor poses a fundamental challenge for NLP, requiring models to integrate world knowledge, pragmatics, and cultural context in order to read -- or write -- between lines. Early computational studies on humor understanding tried to recognize humor by classifying texts to humorous and non-humorous ones \cite{mihalcea2005making}. More recently, the focus has shifted from detecting humor towards rating humor, e.g., asking which of several options is the funniest. For instance, Hessel et al.\ \cite{hessel2023androids} introduced benchmarks derived from the New Yorker Caption Contest and asked LLMs to rank caption funniness as well as to match captions to cartoons. The results show LLMs falling substantially short of human-level judgment in this humor ranking task. This gap was further underscored by Zhang et al.\ \cite{zhang2024humor}: they assembled 250 million crowdsourced human ratings of the New Yorker's cartoon captions and observed systematic divergences between language model and human rankings of funniness of the captions. Directly related to our setting, Ofer and Shahaf \cite{ofer2022cards} introduced a dataset of 300,000 online \textit{Cards Against Humanity} games and trained traditional machine learning models to predict which cards humans select as the winning jokes. 
%Their models performed clearly better than random, but primarily focused on the punchline card only. 
Their models primarily focused on the punchline card only and achieved around 20\% accuracy in card choice prediction. %, an
% Their models achieved accuracies in card choice prediction of around 20\%, and primarily leveraged the punchline card

%Their work established the game as a richly labeled, non-trivial testbed for computational humor research.

\paragraph{\textbf{LLM Alignment}}

LLMs are aligned with human preferences via RLHF \cite{ouyang2022training}, yet this alignment can fail on subjective tasks \cite{ying2025beyond} and skews toward specific demographic groups in ways that resist correction even under explicit steering \cite{santurkar2023whose}. While Argyle et al. \cite{argyle2023out} demonstrate that LLMs can approximate subpopulation response distributions under appropriate conditioning, humor, as a maximally subjective, culturally loaded domain that alignment procedures may not target directly, remains an untested frontier for such demographic mapping.

\section{Methodology}
\label{sec:methodology}
We formalize humor alignment as a discrete preference selection task. Let $\mathcal{M} = \{m_1, \ldots, m_k\}$ denote a set of $k$ LLMs and $\mathcal{G} = \{g_1, \ldots, g_n\}$ a set of $n$ game rounds. Each round $g_i\triangleq (b_i,W_i)$ consists of a black card prompt $b_i$ and a set of candidate white card responses $W_i = \{w_i^1, \ldots, w_i^{|W_i|}\}$. Each model $m_j$ acts as a humor judge, selecting the white card it deems the funniest response:
\begin{equation}
    f_{m_j}(g_i) = w_i^* \in W_i
\end{equation}
To assess preference stability and any position bias, each round is repeated across $R$ replicates, each time following a random permutation $\pi_r$ of the white card order, denoted $f_{m_j}^{(r)}(g_i)$ for replicates $r = 1, \ldots, R$.

\subsection{Datasets}
\paragraph{CAH Lab Gameplay Dataset} This dataset\footnote{Data is available upon request to CAH Lab at
mail@cardsagainsthumanity.com.} consists of games played on the online \textit{Cards Against Humanity} Labs platform\footnote{https://lab.cardsagainsthumanity.com}. Participants engaged with the game voluntarily for entertainment purposes and were not recruited as annotators or crowdworkers for this study. In each round, a player is presented with a prompt card (black card) and ten candidate punchline cards (white cards), and is required to select the punchline they find funniest. The logs were collected between November 2023 and April 2025. More information about the original dataset in Appendix~\ref{app:datasets_gameplay}.

\paragraph{CAH Lab Demographic Answers Dataset} As part of the platform’s optional survey, players can provide demographic information. The dataset contains responses from players, covering six demographic attributes: gender, sexual orientation, race, political ideology, country, and U.S. state. Each entry consists of a player identifier, a demographic category, and the corresponding self-reported answer. The dataset enables analysis of demographic variation in gameplay behavior while preserving anonymity.

\paragraph{Card Topic Labeling} To better interpret players' and LLMs' answer preferences, we annotate all white cards with 1 to 4 topics among 15 possibilities (e.g. ``Anatomy, bodily fluids, gross-out physical humor", ``Sexual content: innuendo, explicit acts, relationships"). The list of possible topics was empirically derived from the distributions of white cards. Cards were annotated using an ``LLM-as-a-judge" scheme using Mixtral 8x7B.  More details on the topic selection strategy and prompting scheme, and the full list of topics, are presented in both  Appendix~\ref{app:datasets_topics} and~\ref{app:datasets_topic-annotation}

%\guillaume{}

%More details on this dataset in Appendix~\ref{app:datasets}.

\subsection{Round Selection}
\label{round_selection}
Each record in the CAH Gameplay dataset captures an online game round where the player picks a white card they think is the funniest given the black card prompt. This includes the black card prompt, the full set of candidate white card responses presented to the player in the round, round completion time, and the identity of the winning card and the identity of the player. To ensure data quality and meaningful human deliberation, we applied three filtering criteria: rounds completed in fewer than 10 seconds were excluded (as likely reflecting inattentive or reflexive judgments), as were rounds exceeding 120 seconds (which may involve distraction or disconnection rather than active play) and rounds explicitly marked as skipped. 

To manage experimental cost while maintaining statistical coverage, we randomly sampled 4,947 rounds from the filtered corpus, preserving the original distribution of black card types and content categories present in the full dataset. More details about this can be found in Appendix~\ref{app:datasets_rounds}.

\subsection{Experimental Setup}
\label{sec:protocol}
We leverage the CAH Lab Gameplay dataset as our human baseline, enabling direct comparison between model selections and the cards that human players actually found funniest under naturalistic conditions.
Each sampled round was presented to 5 LLMs: GPT-5.2, Gemini 3 Flash, Claude Opus 4.5, Grok 4, and DeepSeek-V3.2 as a structured humor selection task. The models were given the black card prompt and a numbered list of ten candidate white card responses, and were instructed to select the single funniest card by responding with its number followed by the exact card text. For rounds requiring two-card combinations, models were additionally instructed to select the funniest card for one blank slot, with \texttt{target\_slot} indicating which blank was being filled. More details on the prompting are given in Appendix~\ref{app:models_prompting}. We also note that the real life CAH game is played where each round, a "Card Czar" reads a black question card, and others submit their funniest white card. The Czar picks the best answer to win a point. Since our dataset doesn't allow for such setup and our purpose is to study humor alignment, we focus on comparing LLM choices to the human players choices. 

Each round was administered across two replicates ($R = 2$), with white cards shuffled into a different random order in each replicate to mitigate position bias in model selections. Responses that could not be resolved to a valid card were recorded as abstentions. Rounds in which any model abstained were flagged as invalid and excluded from agreement analyses, ensuring that all metrics are computed over a consistent set of complete observations across all five models. More details about the models abstentions can be found in Appendix ~\ref{app:models_abstentions}.

\subsection{Humor Benchmarking}
\label{sec:benchmarking}
We study humor alignment along three complementary axes: human--LLM alignment, LLM--LLM alignment, and decision-level analysis. In the latter, we look at the LLM behavior across position bias and topic choice. 

\paragraph{Human--LLM Alignment}
We measure the degree to which the humor judgment of model $m_j$ matches human preference using the accuracy rate:
\begin{equation}
    \text{ACC}(m_j) = \frac{1}{2}\sum_{r=1}^2 \frac{1}{n} \sum_{i=1}^{n} \mathbf{1}\left[ f_{m_j}^{(r)}(g_i) = w_i^\dagger \right],
\end{equation}
where $w_i^\dagger$ denotes the white card picked by the human.

\textit{\textbf{Heterogeneity analysis}} If different socio-demographic groups have different senses of humor, a low average accuracy rate could mask a situation where a model's sense of humor is strongly aligned with some groups' while strongly misaligned with others'. To address this issue and better understand model alignment, we report accuracy rates for different socio-demographic groups, aggregated at the player level across rounds and replicates using the CAH Lab Demographics Answers Dataset. For this analysis, confidence intervals are bootstrapped at the player level.

%As we lack inter-rater agreement measures in the CAH Gameplay dataset

%To assess whether alignment is stable across independent runs, we additionally compute the per-replicate accuracy $\text{ACC}^{(r)}(m_j)$ for each $r \in \{1, \ldots, R\}$ and report 95\% confidence intervals estimated via bootstrapping.

\paragraph{LLM Agreement}
We measure LLM behavior along two dimensions. First, internal consistency quantifies how reliably an LLM reproduces its own judgments across replicates with randomized white card order.

%Second, inter-model pairwise agreement measures the proportion of rounds in which two distinct models select the same response:
%
%\begin{equation}
%    \text{AGR}(m_j, m_l) = \frac{1}{n} \sum_{i=1}^{n} \mathbf{1}\left[ f_{m_j}^{(1)}(g_i) = f_{m_l}^{(2)}(g_i) \right].
%\end{equation}
%

Second, inter-model pairwise agreement measures the proportion of rounds in which two distinct models select the same response across replicates. We define inter-model pairwise agreement between replicates $r$ and $r'$ as:
\begin{equation}
    \text{AGR}^{r, r'}(m_j, m_l) = \frac{1}{n} \sum_{i=1}^{n} \mathbf{1}\left[ f_{m_j}^{(r)}(g_i) = f_{m_l}^{(r')}(g_i) \right].
\end{equation}
We obtain a single measure of inter-model pairwise agreement by averaging $ \text{AGR}^{1, 2}$ and  $\text{AGR}^{2, 1}$.
%\begin{equation}
%\text{AGR}(m_j, m_l) = \frac{1}{2} (AGR^{1, 2}(m_j, m_l) + AGR^{2, 1}(m_j, m_l))
%\end{equation}

%}

%To ensure comparability with $\text{CON}$, agreement is computed cross-replicate: $f_{m_j}$ and $f_{m_l}$ 

%\yousra{modify}

%are taken from different replicates and averaged over both orderings.

Together, the two metrics reveal whether models are self-consistent and whether they agree with each other.

\paragraph{Explaining LLM Humor Behavior} We investigate LLM humor selection along three dimensions. 

\textit{\textbf{Position bias}} We test whether each model's pick distribution across the ten slate positions deviates from uniform using a chi-square goodness-of-fit test:
\begin{equation}
        \chi^2(m_j) = \sum_{p=1}^{|W_i|} \frac{\left(O_p - E_p\right)^2}{E_p}
\end{equation}
where $O_p$ is the number of times model $m_j$ picks position $p$ across all rounds, and $E_p = n / |W_i|$ is the expected count under uniformity, with $n$ the total number of picks by model $m_j$.

\textit{\textbf{Topic bias}} While humor in our setup is intrinsically linked to the match between a white and a black card, LLMs' selections might simply be driven by the topics of the black cards, regardless of context. We conduct an analysis of the distribution of topics present in the different LLMs' answers. To do so, we display a heatmap of the share of each of the 15 white card topics among LLMs' answers. We also display these shares in human picks and among all possible hands of white cards models could pick from for comparison.
% among all possible picks.

\textit{\textbf{Joint position and topic bias analysis}} Finally, we seek to quantify whether the combination of position bias and card topics is sufficient to accurately predict the different LLMs' answers, or if their sense of humor involves more complex mechanisms. To do so, we fit for each model $j$ a conditional logit model, modeling the probability $p_j(k,i)$ of model $j$ picking the card at position $k$ in round $i$ as 
$$ p_j(k,i) = \frac{\exp(x_{ik}^T \beta_j)}{\sum_{k=1}^{|W_i|} \exp(x_{ik}^T \beta_j)}
$$
where $x_{ik}$ describes the $k$-th card at round $i$ in terms of topic flags and (one hot encoded) position, and $\beta_j$ encodes model $j$'s valuation of topics and card position. The models are fitted on 80\% of rounds. We report the round-level accuracy of the learned models on a test set, composed of the remaining 20\% of rounds.

%For \textit{topic and content preferences}, we examine the thematic properties of cards selected by each model and compare humor styles across models to identify systematic content-level differences. 

%\yousra{Will model this accordingly after I get Guillaume's final analysis and results}

%Finally, we conduct a variance analysis to assess the extent to which position bias and topic preference jointly explain the \textit{inter-model agreement} patterns documented in 
%\yousra{Will model this accordingly after I get Guillaume's final analysis and results}
\section{Experiments}
\label{sec:results}
We evaluated the five LLMs on the humor preference selection task. We report results across 4,947 unique rounds with two replicates each; of the resulting 9,894 records, 9,612 passed validation. Unless otherwise noted, all analyses use only valid rounds. More details about the models in Appendix~\ref{app:models}. For brevity, we refer to each model by its family name throughout the following sections (e.g., GPT, Gemini, Claude, DeepSeek, Grok) rather than the exact model version used.

\subsection{Human--LLM Alignment}
\label{sec:accuracy}
\afterpage{
\begin{figure*}[!t]
    \centering
    \includegraphics[width=0.8\linewidth]{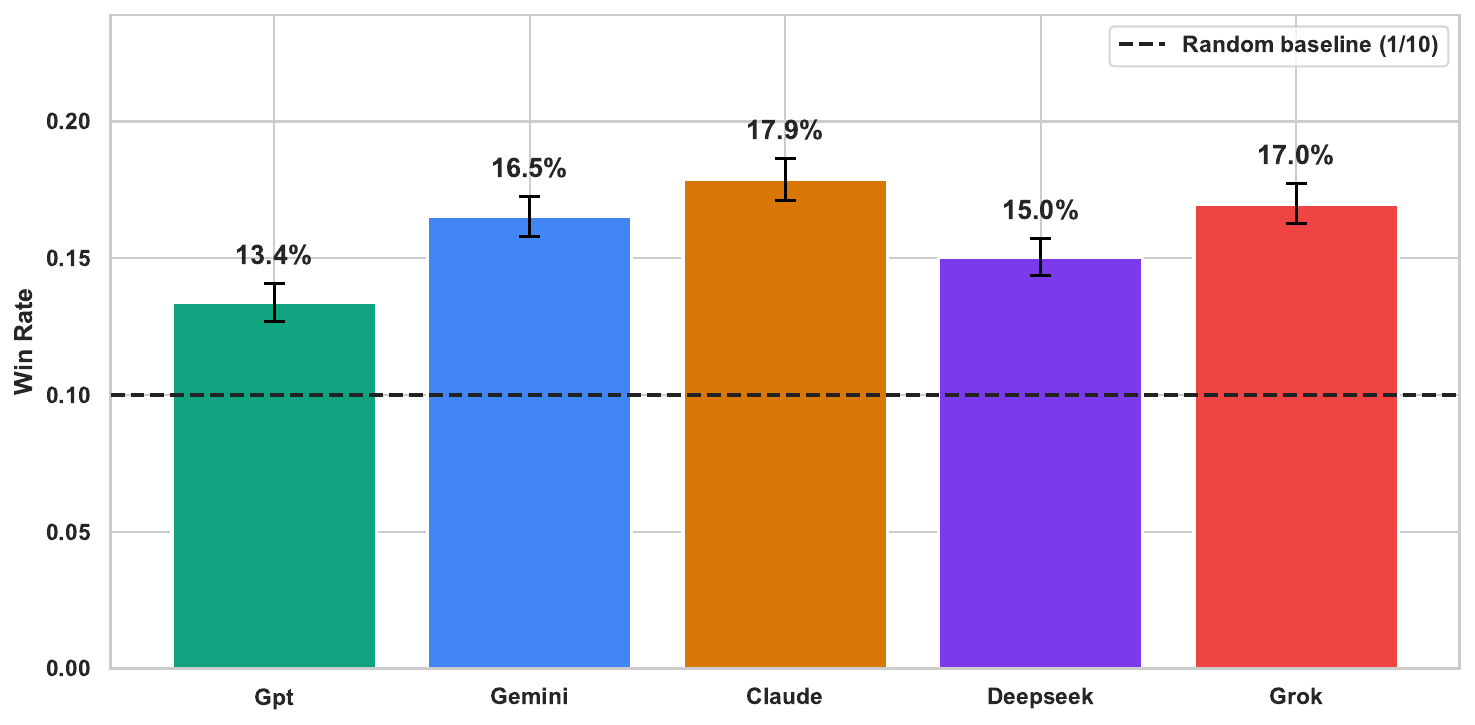}
    \caption{Human-LLM Alignment for all 5 models, with bootstrapped 95\% confidence intervals. The dashed line indicates the random baseline (1/10), reflecting the expected win rate of a model selecting cards uniformly at random from a slate of 10.
    }
    \label{fig:accuracy}
\end{figure*}
}
First, we measure human alignment as the proportion of rounds in which a model selects the same white card as the human-designated winner out of ten candidate cards. While the CAH Lab Gameplay dataset does not allow us to measure inter-annotator agreement, the performance of several baselines can help contextualize these findings. Random card choice would achieve an accuracy of 10\%, whereas picking cards based on popularity would achieve 19.11\% accuracy, and an ensemble of boosted trees learned from human player choices 19.77\%.\footnote{More information on the construction of these baselines is provided in Appendix \ref{app:baseline}. The small gap between the accuracies of the popularity baseline and of the boosted tree ensemble is coherent with the findings of \cite{ofer2022cards}}. 

% XGBoost Top-1 Acc (per slot):  0.1977
%Popularity Baseline:           0.1911

All five models exceed the random baseline of 10\%, as shown in Figure~\ref{fig:accuracy}.
Claude achieves the highest overall alignment, followed by Grok and Gemini, while DeepSeek and GPT  trail behind.
However, absolute alignment rates remain low, ranging between 13\% and 18\%. This suggests that although LLMs capture some aspect of human humor preference, the task remains largely unsolved. 
Performance is consistent across replicates, indicating that the low alignment reflects a genuine limitation rather than model instability.

\textit{\textbf{Heterogeneity analysis}} We turn to the investigation of differences in model alignment across sociodemographic groups covered by the CAH Lab Demographic Answers Dataset. Note that the rounds for which LLM picks were obtained cover 824 players, as 50\% of rounds have missing player IDs and players can play multiple rounds.  Figure \ref{fig:heterogeneity} displays accuracy rates by demographic subgroups at the player level on this population. While we do not have enough statistical power to detect fine-grained demographic differences, the results do not suggest the existence of substantial heterogeneity. In other words, we find no evidence that the low-to-moderate human--LLM alignment found on average is driven by differences in alignment across sociodemographic lines.
\subsection{LLM Agreement}
\label{sec:rq2}
\afterpage{
\begin{figure*}[!t]
    \centering
    \includegraphics[width=0.8\textwidth]{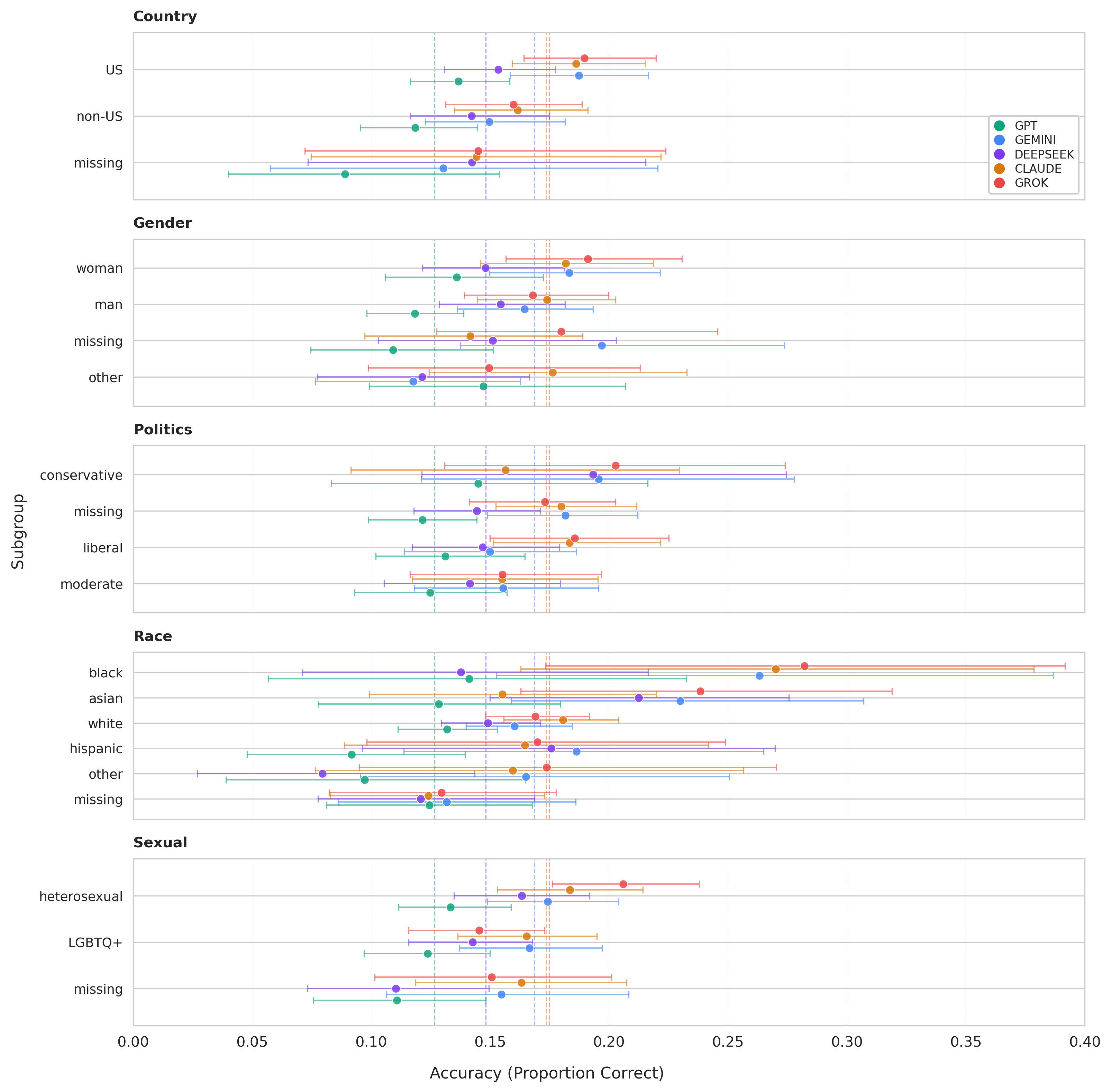}
    \caption{Accuracy rates by demographic subgroup, aggregated at the player level across replicates and rounds. The displayed 95\% confidence intervals are bootstrapped at the player level. Vertical bars indicate player-level mean accuracy of models on the population of the 824 players with non-missing IDs.}
    \label{fig:heterogeneity}
\end{figure*}}
Next, we examine whether models have their own senses of humor, or even converge on a shared `LLM sense of humor', through two complementary lenses: inter-model agreement, and response consistency across replicates.
\begin{figure}[H]
    \includegraphics[width=0.5\textwidth]{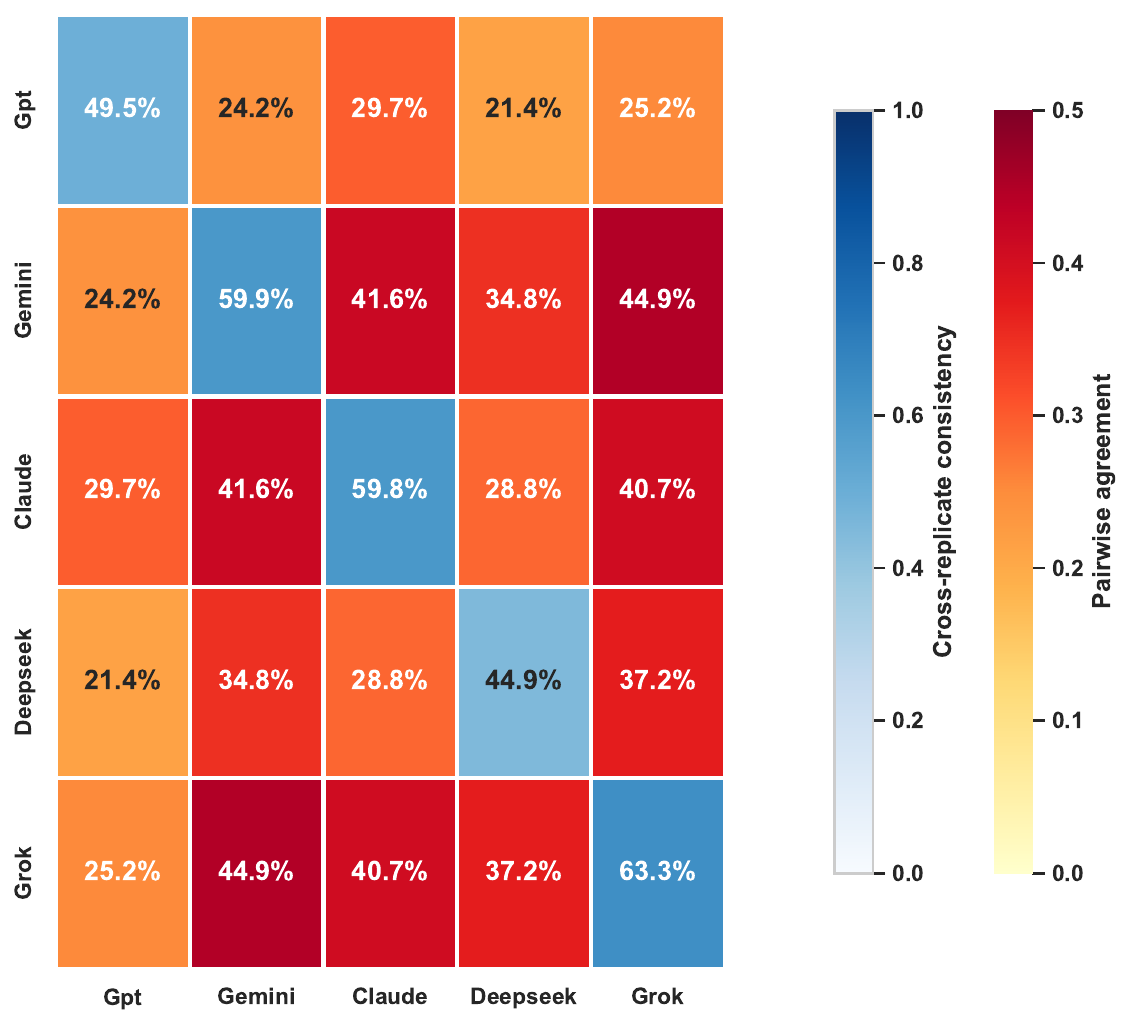}
    \caption{Pairwise agreement rate between models, measured as the proportion of rounds in which two models select the same white card. Off-diagonal cells show inter-model agreement (yellow-red scale); diagonal cells show intra-model consistency across the two replicates (blue scale).
    }
    \label{fig:agreement}
\end{figure}
Figure~\ref{fig:agreement} shows pairwise agreements between LLMs (rows and columns), while the diagonal indicates intra-model consistency, computed as agreement between replicates of the same model. 

Intra-model consistency (cells in the diagonal) is much higher than LLM-human alignment across all models. Grok shows the strongest self-consistency (63.3\%), followed by Gemini (59.9\%) and Claude (59.8\%), while GPT is the least consistent (49.5\%). 
%\hannu{This indicates...randomness, possible position bias, ...?}

Inter-model agreements (other cells than the diagonal) range from 21.4\% to 44.9\%. Thus, remarkably, they also substantially exceed the human-LLM alignment rates of 13–18\% reported in Figure~\ref{fig:accuracy}.

\subsection{Explaining LLM Humour Behavior}

Next we analyze two types of possible biases in LLMs in this task: position bias and topic bias.

\subsubsection{Position bias}

\begin{table}
  \centering
  \begin{tabular}{lrrl}
    \hline
    \textbf{Model} & \textbf{$\chi^2$} & \textbf{\textit{p}-value} & \textbf{Dominant Position} \\
    \hline
    GPT     & 356    & $<$0.001   & —     \\
    Gemini   & 282  & $<$0.001 & — \\
Claude   & 678  & $<$0.001 & Early positions \\
DeepSeek & 1851 & $<$0.001 & Position 3 \\
Grok     & 658  & $<$0.001 & Position 10  \\\hline
  \end{tabular}
  
  \caption{Chi-square test of uniformity for position bias per model (df=9). Higher $\chi^2$ indicates stronger deviation from a uniform pick distribution across the 10 slate positions.}
  \label{tab:chitest}
\end{table}

All five models exhibit significant deviation from a uniform pick distribution as displayed in Table~\ref{tab:chitest}. This aligns with finding from previous literature \cite{pezeshkpour2024large}. DeepSeek shows the strongest position bias, driven by a pronounced concentration of picks at Position~3. Claude and Grok display similarly strong bias, with Grok favoring the last position and Claude showing moderate spread with a mild early-position preference. GPT and Gemini exhibit the weakest bias, though still highly significant, with picks distributed more evenly across the slate. We show this in Appendix ~\ref{app:explaining}.
%\hannu{...this partially explains modest human-llm agreements and intra-model consistencies...}

\subsubsection{Topic bias}

\begin{figure*}[]
    \centering
    \includegraphics[width=\linewidth]{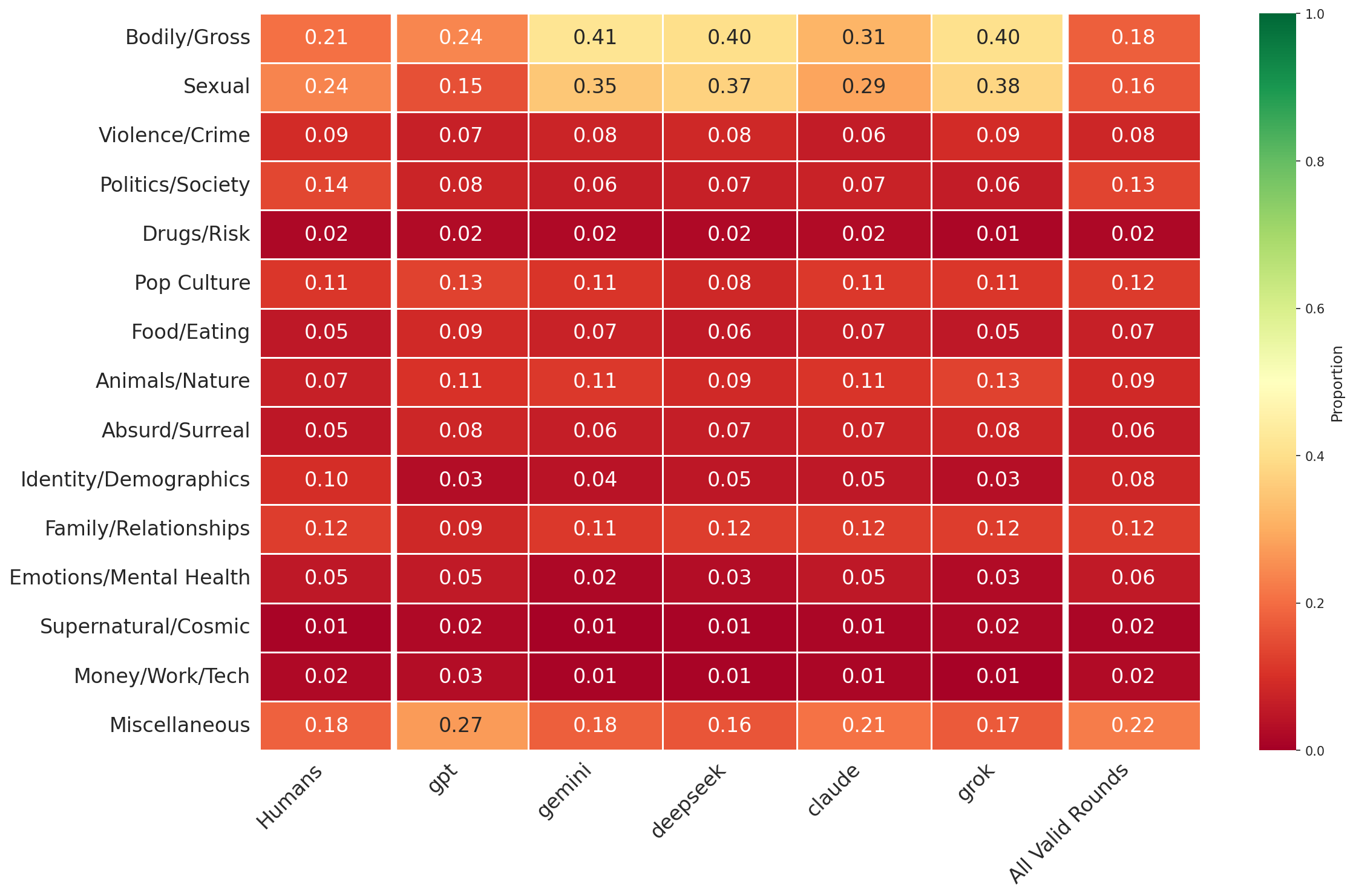}
    \caption{Shares of human (column 1) and LLM (columns 2-5) white card picks involving different topics (a card can have several topic tags). Column 6 provides the shares of topics among all available cards for reference. 
    }
    \label{fig:topic_heatmap}
\end{figure*}

We turn to the description of the topics present in the white cards chosen by LLMs, displayed in Figure \ref{fig:topic_heatmap}. In coherence with pairwise agreement patterns, we find commonalities between the topics involved in Gemini, DeepSeek, Claude and Grok's responses. In particular, these models' card picks often rely on bodily humor (31\% to 40\% of answers, compared to 21\% for humans) and sexual themes (29\% to 38\%, against 24\% for humans). On the other hand, GPT makes less use of cards with these topics (24\% and 15\% respectively), and stands out in terms of the use of ``miscellaneous"-themed cards (27\% of answers, against 16\% to 18\% for other models and humans). It is also worth noting that all models' responses involve fewer answers related to ``politics/society" and ``identity/demographic" topics (6-8\% and 3-5\% of answers respectively) than humans (14\% and 10\%).

\subsubsection{Joint position and topic bias analysis}

Finally, we assess to what extent the combination of position and topic biases predicts LLMs' responses. We fit conditional logistic models with card position and topic as covariates to predict LLM behavior based on the list of white answer cards only, ignoring the black card \cite{ofer2022cards}. Figure \ref{tab:surrogate_accuracies} displays the share of rounds in the test set for which logistic models correctly predict LLMs' answers. Note that an upper bound for such a prediction quality metric is given by the LLMs' own self-consistency, whereas if both topics and position biases were irrelevant, the metric would be 0.1. The surrogate models for Grok, DeepSeek and Gemini correctly characterize chosen white cards for more than a third of rounds (0.361, 0.351 and 0.339). %For DeepSeek in particular, the combination of a surrogate logistic model accuracy of 0.351 with the LLM's own self-consistency rate of 44.9\% suggests that a large part of its responses can be explained by randomness, white card topic regardless of the black card prompt, and position bias. 
The surrogate conditional logistic models for GPT and Claude achieve markedly lower round-level accuracies, at 0.171 and 0.244. Altogether, these results suggest that a non-trivial share of model responses can be explained by position bias and white card content preference, but also suggests  the existence of more complex mechanisms at play in LLM humor.

\begin{table}[h]
\centering
\begin{tabular}{lc}
\hline
Model & Round-level accuracy \\
\hline
GPT & 0.171 \\
Gemini & 0.339 \\
Claude & 0.244 \\
DeepSeek & 0.351 \\
Grok & 0.361 \\
\hline
\end{tabular}
\caption{Share of rounds for which surrogate logistic models correctly predict LLMs' answers.}
\label{tab:surrogate_accuracies}
\end{table}

\section{Discussion}
%The accuracy ceiling in the human-llm alignment likely reflects both the inherent subjectivity of humor and the fact that CAH provides a space for, and even rewards, transgressive incongruity. This mode of humor may conflict with the safety-oriented fine-tuning of current LLMs. That all models consistently outperform the baseline confirms that LLMs do capture meaningful aspects of human humor preferences; however, \textbf{we hypothesize that the modest accuracy indicates that they rely on shallow heuristics rather than deeper pragmatic reasoning that drives human humor selection}.

%We also find that \textbf{models agree with each other far more than they agree with humans}. It indicates that LLMs have converged towards a shared but human-misaligned notion of humor. 
%since they are consistent among themselves, and even self-consistent across runs. However, this internal consensus does not translate into better prediction of human preferences. 
%This suggests a systematic bias in how frontier LLMs process humor of the type in CAH. This likely reflects shared patterns in instruction tuning or RLHF that reward certain types of responses over the transgressive and absurdist style that makes CAH funny to humans.

Altogether, our analysis of the \textit{Cards Against Humanity} gameplay of LLMs points towards the existence of i) an accuracy ceiling in current human-LLM humor alignment, and ii) the convergence of frontier LLMs towards a somewhat shared, but human-misaligned, notion of humor.

Our findings suggest LLMs \textbf{capture some meaningful aspects of human humor preferences}: accuracy rates of LLMs significantly outperform a random baseline in terms of human-LLM agreement. Nevertheless, these accuracy rates are only moderate (13-18\%), whereas simple baselines, as presented in Section \ref{sec:accuracy}, suggest accuracies around at least 19\%-20\% can be achieved. 
This average finding does not seem to be driven by underlying considerable differences in alignment to different demographic subgroups.

Moreover, we also find that LLMs \textbf{agree with each other far more than they agree with humans}. This may indicate that LLMs have converged to some extent towards a shared but human-misaligned notion of humor where bodily and sexual topics are emphasized more, and demographic topics less, than by humans.

\textbf{We hypothesize that LLMs rely partly on shallow heuristics rather than deeper pragmatic and contextual reasoning that drives human humor selection}. Indeed, we find models to exhibit significant position bias and commonalities in topic choices in their answers, as shown by surrogate model accuracies up to 36\%.

%and a non-trivial part of their choices to be predictable based on position and card topics, regardless of context.

The current accuracy ceiling in human-LLM humor alignment also likely reflects both the inherent subjectivity of humor and the fact that CAH provides a space for, and even rewards, transgressive incongruity. This mode of humor \textbf{may conflict with the safety-oriented fine-tuning of current LLMs}: shared patterns in instruction tuning or RLHF that reward certain types of responses over the transgressive and absurdist style that makes CAH funny to humans. Consistent with this hypothesis, the analysis of topic choices reveals LLMs to choose white cards about "politics/society" and "identity/demographics" topics less frequently than humans, or compared to the distribution of such cards among possible answers.

\section{Conclusion and Future Work}

We presented a large-scale study of humor alignment between frontier language models and humans, using \textit{Cards Against Humanity} as a naturalistic preference selection testbed. Across 9,894 rounds and five models, 
%we find that LLMs consistently exceed the random baseline but only achieve modest alignment with human humor judgment.
we find LLMs' alignment with human humor judgments to be modest.
Strikingly, models agree with each other substantially more than they agree with humans. Further analysis reveals that this divergence is partly explained by systematic position biases and content preferences, raising important questions about the extent to which LLM behavior on tasks involving humor reflects genuine understanding or structural artifacts of inference.

Several directions emerge naturally from this study. A larger experimental scale — more rounds, replicates, and a more culturally diverse set of models — would yield more robust estimates and enable cross-cultural comparison. Extending the framework to other humor formats such as memes or satirical writing would test the generalizability of our findings. Finally, our observation that models agree with each other more than with humans raises the question of whether specific human subgroups, defined by humor style rather than demographics, show substantially higher LLM alignment.

%These findings have broader implications for LLM evaluation and alignment research. Humor, as one of the most culturally embedded and subjective dimensions of human communication, exposes limitations that standard benchmarks do not. The gap between inter-model agreement and human alignment suggests that models may be converging on a shared but impoverished notion of funniness --- one that is internally consistent yet disconnected from the social and contextual factors that make humor meaningful to humans. We hope this work encourages the community to look beyond aggregate accuracy metrics and investigate the mechanisms underlying model preferences on subjective tasks.

\section{Limitations}

\begin{itemize}
    %\item \textbf{Prompt sensitivity.} Prompt design plays a critical role in model performance. LLMs are sensitive to the phrasing and structure of inputs, and small variations can lead to substantially different outputs. This dependence on careful prompt engineering limits the scalability and generalizability of the approach to new tasks and domains.

    \item \textbf{Experimental scale.} While our study covers 9,894 rounds across two replicates, a larger number of rounds and replicates would yield more stable estimates and stronger statistical conclusions. Expanding the experimental scale was constrained by the API costs associated with querying five frontier models across multiple runs.
    \item \textbf{Single-player ground truth.} The CAH Lab dataset provides the choice of a single player for each round, which we use as human reference. However, the data does not allow us to estimate inter-rater agreement, which would help contextualize the modest LLM accuracy rates, and distinguish model failure from inherent task subjectivity.
    \item \textbf{Temperature Setting.} All models were queried at a fixed temperature of 0.8, allowing for some response variability while maintaining comparability across models. Varying this parameter could yield different self-consistency profiles and is left for future investigation.
 \item \textbf{LLM-as-judge for topic labeling.} Our content analysis relies on LLM-generated labels to categorize white card themes, introducing potential noise and systematic bias. We manually verified label coherence on a subset of cards, finding the labels to be generally consistent and meaningful. However, a more systematic validation methodology such as inter-annotator agreement with human raters would strengthen the reliability of this analysis and is left as future work.

    \item \textbf{Player population.} The dataset reflects the preferences of a specific population of CAH players, who are predominantly Western and self-selected. Broader coverage of player demographics and cultural backgrounds would be needed to generalize our human alignment findings.
    
    \item \textbf{Western model bias.} Four of the five models --- GPT, Gemini, Claude, and Grok --- are developed primarily in a Western context, and their humor profiles may reflect cultural biases embedded in their training data and alignment procedures. DeepSeek is the only exception, though a single non-Western model is insufficient to draw cross-cultural conclusions.
\end{itemize}

\section*{Acknowledgments}
We would like to thank our colleagues MaryBeth Defrance, Edith Heiter and Jefrey Lijffijt for their valuable comments and feedback. The research leading to these results was funded/co-funded by the European Union (ERC, VIGILIA, 101142229), the Special Research Fund (BOF) of Ghent University (BOF20/IBF/117), the Flemish Government under the ``Onderzoeksprogramma Artificiële Intelligentie (AI) Vlaanderen'' programme, and the FWO (project no. G073924N). Views and opinions expressed are however those of the author(s) only and do not necessarily reflect those of the European Union or the European Research Council Executive Agency. Neither the European Union nor the granting authority can be held responsible for them. For the purpose of Open Access the author has applied a CC BY public copyright license to any Author Accepted Manuscript version arising from this submission.

\bibliography{custom}

\appendix

\include{appendix}
\end{document}

%% file: appendix.tex
\section{Datasets Details}
\label{app:datasets}

\subsection{CAH Gameplay dataset}
\label{app:datasets_gameplay}

The raw dataset had 148,497 past games (rounds). There are 501 unique black prompt cards and 2074 white punchline cards, resulting in 1,484,970 possible unique jokes (where a joke is the result of filling in the blank of the
prompt card with a punchline).

%filled out their informations in optional survey and (i.e: they appear in the CAH Lab Demographic Answers Dataset)

\subsection{Topic list selection}
\label{app:datasets_topics}

To obtain the list of 15 topics used for the analysis, we prompted a version of GPT-5 accessed through Microsoft Copilot to provide a taxonomy of topics based on the following prompt: \textit{``The following short pieces of text are "white cards" in the game "Card against humanity". Help me find a taxonomy of topics they cover, so I can label them downstream."}, followed by white cards. This prompt was run twice, with two different samples of 100 white cards, yielding two different taxonomies which were then harmonized.

Table \ref{tab:topic_labels} provides the resulting list of topics (with their short-hand labels, as used in Figure \ref{fig:topic_heatmap}).

\begin{table}[htbp] \small %\scriptsize
\centering
\caption{Topic Labels and Definitions}
\label{tab:topic_labels}
\begin{tabular}{@{}lp{4.5cm}@{}}
%\begin{tabular}{|l|c|}

\toprule
\textbf{Label} & \textbf{Definition} \\
\midrule
Bodily/Gross & Anatomy, bodily fluids, gross-out physical humor \\
Sexual & Sexual content: innuendo, explicit acts, relationships \\
Violence/Crime & Physical harm, mortality, criminal acts, threats \\
Politics/Society & Government, activism, social norms, cultural commentary \\
Drugs/Risk & Substance use, addiction, reckless actions \\
Pop Culture & Celebrities, movies, memes, brands, viral trends \\
Food/Eating & Meals, ingredients, dining, consumption \\
Animals/Nature & Wildlife, pets, ecosystems, biological refs \\
Absurd/Surreal & Illogical juxtapositions, nonsense, anti-humor \\
Identity/Demographics & Race, gender, age, disability, sexuality, nationality \\
Family/Relationships & Parenting, friendships, domestic life, mundane interactions \\
Emotions/Mental Health & Anxiety, joy, depression, coping, psychological framing \\
Supernatural/Cosmic & Ghosts, aliens, magic, existential cosmic themes \\
Money/Work/Tech & Jobs, finance, digital life, institutional critique \\
Miscellaneous & Concrete items/concepts not captured above \\
\bottomrule
\end{tabular}
\end{table}

\subsection{Topic annotation prompt}
\label{app:datasets_topic-annotation}

White cards were annotated with topics using the following prompt, passed to Mixtral 8x7B.

\begin{tcolorbox}[
  colback=gray!8, colframe=gray!40,
  fontupper=\ttfamily\small,
  boxrule=0.4pt, left=6pt, right=6pt, top=4pt, bottom=4pt,
  breakable,
  width=\linewidth,
  before upper={\setlength{\emergencystretch}{3em}\sloppy}
]
%\begin{tcolorbox}[colback=gray!8, colframe=gray!40, %fontupper=\ttfamily\small, 
%boxrule=0.4pt, left=6pt, right=6pt, top=4pt, bottom=4pt]

'''
You are an expert annotator labeling "white cards" from Cards Against Humanity for humor research.

\#\#\# TASK
Analyze the input card and return a JSON object matching the schema below.\\

\#\#\# OUTPUT SCHEMA (STRICT)\\
\{\\
  "topics": ["slug1", "slug2"],           // 1-4 items from TOPICS list\\
\} \\
\\
\#\#\# TOPICS (use slugs exactly; select 1-4) \\
1. bodily\_functions\_gross\_out         // Anatomy, bodily fluids, gross-out physical humor
2. sexual\_themes                      // Sexual content: innuendo, explicit acts, relationships\\
3. violence\_crime\_death\_threat        // Physical harm, mortality, criminal acts, threats\\
4. politics\_ideology\_society\_culture  // Government, activism, social norms, cultural commentary\\
5. drugs\_alcohol\_risky\_behavior       // Substance use, addiction, reckless actions\\
6. pop\_culture\_media\_consumerism      // Celebrities, movies, memes, brands, viral trends\\
7. food\_eating\_consumables            // Meals, ingredients, dining, consumption\\
8. animals\_nature\_creatures           // Wildlife, pets, ecosystems, biological refs\\
9. absurdism\_surreal\_nonsensical      // Illogical juxtapositions, nonsense, anti-humor\\
10. identity\_demographics\_traits      // Race, gender, age, disability, sexuality, nationality\\
11. family\_relationships\_everyday     // Parenting, friendships, domestic life, mundane interactions\\
12. emotional\_states\_mental\_health    // Anxiety, joy, depression, coping, psychological framing\\
13. supernatural\_cosmic\_paranormal    // Ghosts, aliens, magic, existential cosmic themes\\
14. money\_work\_technology\_modern      // Jobs, finance, digital life, institutional critique\\
15. random\_objects\_miscellaneous      // Concrete items/concepts not captured above \\

\#\#\# RULES \\
1. Use slugs exactly as written (underscores, lowercase, no spaces).\\
2. Output ONLY valid JSON. No markdown, no preamble, no extra text. \\
'''

USER\_PROMPT\_TEMPLATE = ''' \\
\#\#\# INPUT CARD\\
"\{card\_text\}"\\

\#\#\# OUTPUT
'''

\end{tcolorbox}

All 2074 unique white cards in the gameplay dataset were successfully annotated with 1 to 3 topics. Coherence of the annotations was validated by the authors. We also note the topic annotation was done on our final dataset that we've gotten after the round selection.

%\guillaume{TODO ADD EXAMPLES}

%More details about Humour Types (Guillaume)
\subsection{Round Selection Details}
\label{app:datasets_rounds}
After applying the filtering criteria mentioned in Section \ref{round_selection}, we obtain a dataset of 107,562 unique rounds. This dataset contains 493 unique black cards and 2,072 unique white cards, resulting in 1,075,620 possible unique jokes. After randomly sampling rounds from this filtered dataset, we get 4,947 rounds.
%We map the rounds to the players and we find that we have \guillaume ??? players, 824 of whom have a player ID.
Among these rounds, 2,437 can be mapped to a player with valid identifier; there are 824 such players in the studied dataset.

\section{Models}
\label{app:models}
We evaluate five conversational large language models (LLMs) in our experiments: 
\begin{itemize}
    \item \texttt{gpt-5.2} (OpenAI) \cite{singh2025openai}.
    \item \texttt{gemini-3-flash-preview} (Google) \cite{doshi2025gemini3flash}.
    \item \texttt{claude-opus-4-5} (Anthropic) \cite{anthropic2025claudeopus45}.
    \item \texttt{grok-4} (xAI) \cite{xai2025grok4}.
    \item \texttt{DeepSeek-V3.2} (DeepSeek) \cite{liu2025deepseek}.
\end{itemize}
This selection reflects the current landscape of frontier LLMs, covering both open-weight and proprietary systems trained on diverse corpora with differing alignment objectives. All models were queried via their respective APIs with a fixed temperature of 0.8, balancing response diversity with reproducibility, and a maximum output length of 128 tokens, sufficient for the card selection format required. To isolate humor preference from instruction-following variability, all models received an identical prompt template, described in Appendix~\ref{app:models_prompting}. 
All outputs were collected between the 17th of January and the the 10th of March 2026.

\subsection{Model Prompting}
\label{app:models_prompting}

All models received the same prompt template, instantiated from the black card and white card slate. No examples or chain-of-thought instructions were provided, so as to elicit each model's intrinsic humor judgment. Models were instructed to respond with only the number and exact text of their chosen card to facilitate automated parsing.
\subsubsection{Single-Blank Prompt}

For black cards with a single blank, models were asked to select the funniest card from their hand:

\begin{tcolorbox}[colback=gray!8, colframe=gray!40, fontupper=\ttfamily\small, boxrule=0.4pt, left=6pt, right=6pt, top=4pt, bottom=4pt]
You are playing Cards Against Humanity. It's just a game so it's okay to use profanity and to be politically incorrect.\\
Choose the single funniest white card.\\
\\
Black card:\\
\{black\_card\}\\
\\
Your hand:\\
1. \{card\_1\}\\
2. \{card\_2\}\\
\ldots\\
\\
Respond ONLY with:\\
<number>. <exact card text>
\end{tcolorbox}

\noindent where \texttt{\{black\_card\}} is the text of the black card with the blank represented as an underscore, and \texttt{\{card\_1\}\ldots\{card\_N\}} are the white cards in the slate ($N=10$ in all rounds).

\subsubsection{Multi-Blank Prompt}

For black cards with two blanks, a target slot was designated (BLANK \#1 or BLANK \#2) and models were asked to fill only that slot. This keeps the response format consistent across all rounds and allows positional effects to be analyzed independently. It does not capture combinatorial humor arising from the interaction between two cards.

\begin{tcolorbox}[colback=gray!8, colframe=gray!40, fontupper=\ttfamily\small, boxrule=0.4pt, left=6pt, right=6pt, top=4pt, bottom=4pt]
You are playing Cards Against Humanity. It's just a game so it's okay to use profanity and to be politically incorrect.\\
Choose the funniest white card to fill BLANK \#\{target\_slot\}.\\
\\
Black card:\\
\{black\_card\}\\
\\
Your hand:\\
1. \{card\_1\}\\
2. \{card\_2\}\\
\ldots\\
\\
Respond ONLY with:\\
<number>. <exact card text>
\end{tcolorbox}

\noindent where \texttt{\{black\_card\}} is the text of the black card with blanks represented as underscores, \texttt{\{target\_slot\}} is either 1 or 2 indicating which blank to fill, and \texttt{\{card\_1\}\ldots\{card\_N\}} are the white cards in the slate ($N=10$ in all rounds).
%\section{Human-LLM Humour Alignment Details}
%\yousra{I need to chage this because it's for the old dataset}
%\paragraph{Single-Blank vs.\ Multi-Blank Prompts.}

%The dataset contains 7,807 single-blank and 939 multi-blank round-replicates (prompts requiring a card for a specific slot among multiple blanks).
%Most models show a performance drop on multi-blank prompts.
%Grok exhibits the steepest decline, from 18.0\% on single-blank to 12.2\% on multi-blank rounds.
%Claude is the most robust, declining only from 18.0\% to 16.4\%.
%GPT shows virtually no difference (13.4\% vs.\ 13.3\%), though this may reflect its generally lower baseline rather than true robustness to multi-blank complexity.
%{\color{red}Here we have all the sociodemographics additional analyses}
%\subsection{LLM-LLM Alignment Details}
%{\color{red}Any other results for LLM agreements besides the correlation matrix}
\subsection{Models Abstentions}
\label{app:models_abstentions}
Not all models engaged with every round. Gemini exhibited a notably higher abstention rate, producing null or failed picks in 280 rounds (2.8\% of its total records), likely reflecting its content moderation filters triggering on CAH's characteristically offensive prompts. All other models showed near-zero failure rates: GPT, Claude, and DeepSeek recorded no null picks whatsoever. Grok produced only 2 failed picks. One of the fails was caused by a connection error rather than a content refusal. These abstentions were excluded from downstream analyses to ensure that win rate comparisons reflect genuine card selection behavior rather than differential willingness to engage with the game's content.

\section{Explaining LLM Humor Behavior Details}
\label{app:explaining}
\begin{figure*}[t]
    \centering
    \includegraphics[width=\textwidth]{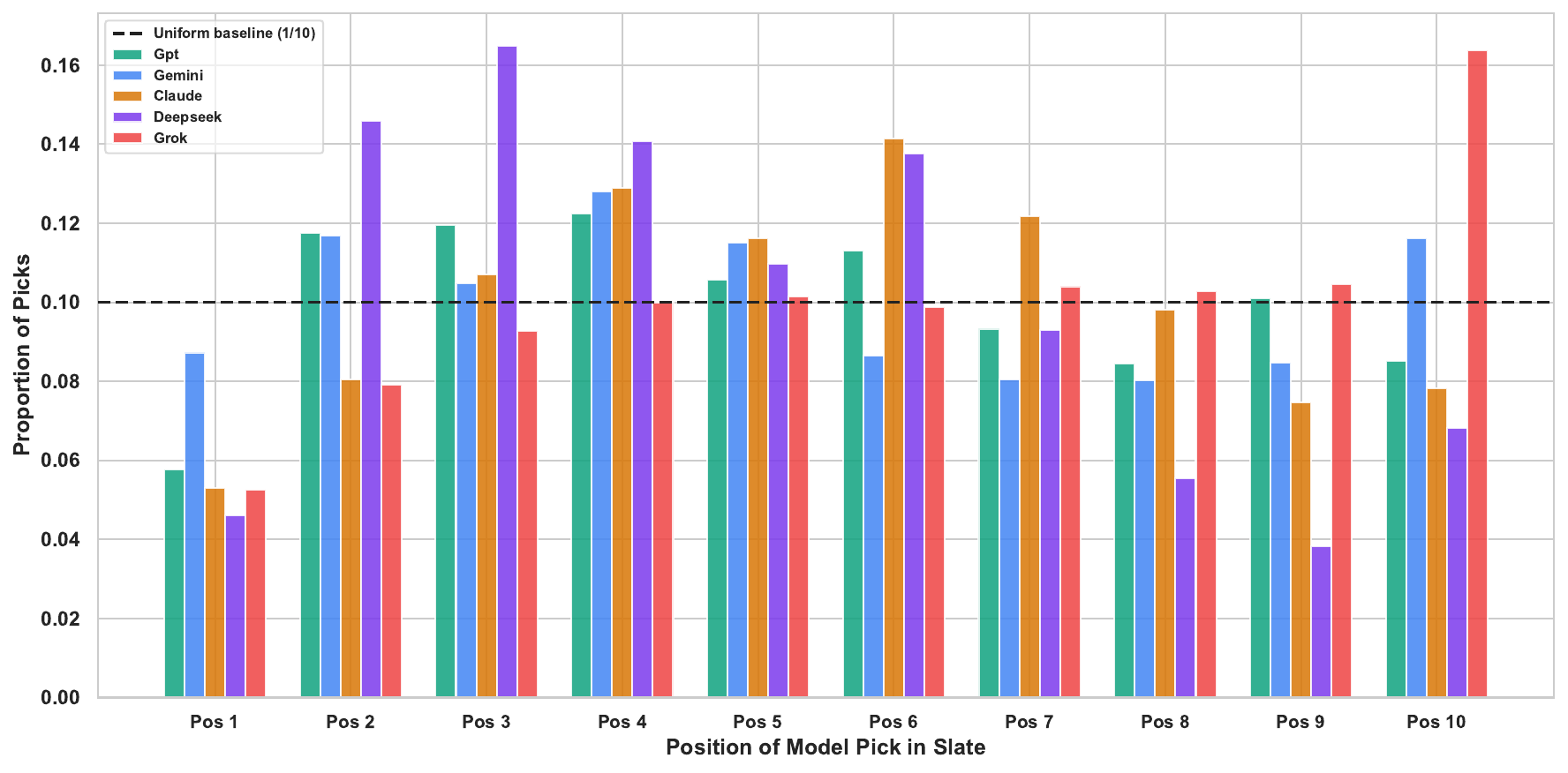}
    \caption{Distribution of model picks across slate positions (N=10). Each bar 
    represents the proportion of rounds in which a model selected the card at a given 
    position, aggregated across all rounds and replicates. Under the null hypothesis 
    of no positional bias, picks would be uniformly distributed at 0.10 (dashed line).}
    \label{app:fig:position}
\end{figure*}

We showcase the distribution of model picks across the slate positions in Figure ~\ref{app:fig:position}. All five models deviate from the uniform baseline, confirming the presence of positional bias across the board.
%\begin{figure}[h]
   % \center
  %  \includegraphics[width=\textwidth]{latex/images/position_bias.pdf}
 %   \caption{Distribution of model picks across slate positions (N=10). Each bar represents the proportion of rounds in which a model selected the card at a given position, aggregated across all rounds and replicates. Under the null hypothesis of no positional bias, picks would be uniformly distributed at 0.10 (dashed line). 
%    }
%    \label{app:fig:position}
%\end{figure}

\section{Baselines predicting human behavior}
\label{app:baseline}

In the absence of proper inter-annotator agreement measures in the CAH Lab Gameplay dataset, we construct two simple baselines (a card popularity baseline, and boosted tree predictions of human choices) to better contextualize the accuracy rates achieved by different LLMs, building on previous work exploration of human humor prediction in CAH \cite{ofer2022cards}. Both baselines are trained on a training set of 84,110 CAH rounds, obtained by removing from the CAH Gameplay dataset all LLM-annotated rounds, and data from any players who participated in these rounds (when player IDs were known).

The first baseline predicts white card choices based on their winrates in a training set (disregarding associated black cards). Despite its simplicity, this strategy was noted to be a strong baseline \cite{ofer2022cards}, outperforming more elaborate modeling attempts in predicting human choices when card position information is not available, as is the case in the dataset. 

The second baseline trains an XGBoost classifier on human choices to predict winning white cards in a given round. The learning task is formulated as a binary classification objective at the card-option level. Each row corresponds to a white card in a given round, with a binary label $y \in \{0, 1\}$ indicating if it was picked that round. Features describing the card-round combination are: i) a 384-dimensional embedding of the white card text; ii) a 384-dimensional embedding of the black card text; iii) the 15-topic annotations for the white card, leading to a total of 783 features. White and black card embeddings are obtained using a Sentence-Bert embedding of the card texts using \textit{all-MiniLM-L6-v2}. To compute test accuracies, we simply compare the card with highest predicted pick probabilities in the round to the actual human choice. Hyperparameters are tuned via \textit{scikit-learn}'s \textit{RandomizedSearchCV} (50 iterations, 3-fold stratified CV) optimizing ROC-AUC; considered options are listed in Table \ref{tab:xgb_hyperparams}. The final model is refit on the full training set.

\begin{table}[h]
\centering
\caption{Hyperparameter Search Space for XGBoost Baseline}
\label{tab:xgb_hyperparams}
\begin{tabular}{ll}
\toprule
\textbf{Hyperparameter} & \textbf{Search Values} \\
\midrule
\textit{n\_estimators} & 100,150,200,250,300 \\
\textit{max\_depth} & 4,5,6,7,8 \\
\textit{learning\_rate} & 0.05,0.08,0.1,0.15,0.2 \\
\textit{min\_child\_weight} & 3,5,7,10\\
\textit{scale\_pos\_weight} & 7,8,9,10,11 \\
\textit{subsample} & 0.7,0.8,0.9,1.0 \\
\textit{colsample\_bytree} & 0.7,0.8,0.9,1.0 \\
\bottomrule
\end{tabular}
\end{table}